\documentclass[runningheads]{llncs}
\usepackage{graphicx}
\usepackage{amsmath}
\usepackage{amssymb}
\usepackage{algorithm}
\usepackage{algpseudocode}

\algnewcommand\And{\textbf{and}}
\algnewcommand\Or{\textbf{or}}
\usepackage{booktabs}
\usepackage{multirow}
\usepackage[bordercolor=gray!20,backgroundcolor=blue!10,linecolor=red,textsize=footnotesize]{todonotes}
\newcommand{\mytodo}[2][]{{%
 \let\marginpar\marginnote
 \reversemarginpar
 \renewcommand{\baselinestretch}{0.8}%
 \todo[#1]{#2}}}
\begin{document}
\title{MG-NET: Leveraging Pseudo-Imaging for Multi-Modal Metagenome Analysis
}
\titlerunning{MG-NET: Multi-Modal Metagenome Analysis}
%
\author{Sathyanarayanan N. Aakur\inst{1} \and
Sai Narayanan \inst{2} \and
Vineela Indla \inst{1} \and
Arunkumar Bagavathi \inst{1} \and
Vishalini Laguduva Ramnath\inst{1} \and
Akhilesh Ramachandran \inst{2}}
\authorrunning{SN. Aakur et al.}
%
\institute{Department of Computer Science, Oklahoma State University, Stillwater, OK, USA \and
Oklahoma Animal Disease Diagnostic Laboratory, College of Veterinary Medicine, Oklahoma State University, Stillwater, OK, USA \\
\email{\{saakurn, ssankar, vindla, abagava, vlagudu, rakhile\}@okstate.edu}\\}
\maketitle              
\begin{abstract}
The emergence of novel pathogens and zoonotic diseases like the SARS-CoV-2 have underlined the need for developing novel diagnosis and intervention pipelines that can learn rapidly from small amounts of labeled data. Combined with technological advances in next-generation sequencing, metagenome-based diagnostic tools hold much promise to revolutionize rapid point-of-care diagnosis. However, there are significant challenges in developing such an approach, the chief among which is to learn self-supervised representations that can help detect novel pathogen signatures with very low amounts of labeled data. This is particularly a difficult task given that closely related pathogens can share more than $90\%$ of their genome structure. 
In this work, we address these challenges by proposing MG-Net, a self-supervised representation learning framework that leverages multi-modal context using pseudo-imaging data derived from clinical metagenome sequences. We show that the proposed framework can learn robust representations from \textit{unlabeled data} that can be used for downstream tasks such as metagenome sequence classification with limited access to labeled data. Extensive experiments show that the learned features outperform current baseline metagenome representations, given only $1000$ samples per class. 

\keywords{Metagenome Analysis  \and Automatic Diagnosis with Metagenomics \and Multi-modal disease intervention}
\end{abstract}
\section{Introduction}
Advances in DNA sequencing technologies~\cite{metzker2010sequencing,mikheyev2014first} have made possible the determination of whole-genome sequences of simple unicellular (e.g., bacteria) and complex multicellular (e.g., human) organisms at a cheaper, faster, and larger scale.
The abundance of collected genome sequences require reliable and scalable frameworks to detect novel pathogens and study further mutations of such pathogens to mitigate threatening disease transmissions. 
Zoonotic diseases, like SARS-CoV-2, are a prime example for the need for rapid learning from noisy and limited data, due to their ability to mutate and cause pandemic situations. 
To this end, DNA sequencing-based approaches, such as metagenomics, have been explored by several researchers~\cite{chiu2019clinical,hasan2020metagenomics} for plant and animal disease diagnostics.
Metagenome-based diagnostics are pathogen agnostic and theoretically have unlimited multiplexing capability. Unlike traditional methods, metagenome-based diagnostics can also provide information on the host's genetic makeup that can aid in personalized medicine~\cite{hamburg2010path}.  
However, metagenome diagnostics encounter the problem of long-tail distribution of pathogen sequences in the data. The problem aggravates for pathogen detection tasks when we consider pathogens from the same genus. For example, \textit{Mannheimia haemolytica} and \textit{Pasteurella multocida} share as much as $95.5\%$ of their genome.
In this work, we consider the Bovine Respiratory Disease Complex (BRD) as a model and aim to detect the presence of six associated bacterial pathogens, namely \textit{Mannheimia haemolytica}, \textit{Pasteurella multocida}, \textit{Bibersteinia trehalosi}, \textit{Histophilus somni}, \textit{Mycoplasma bovis}, and \textit{Trueperella pyogenes}. 

\begin{figure*}[t]
    \centering
    \includegraphics[width=0.99\textwidth]{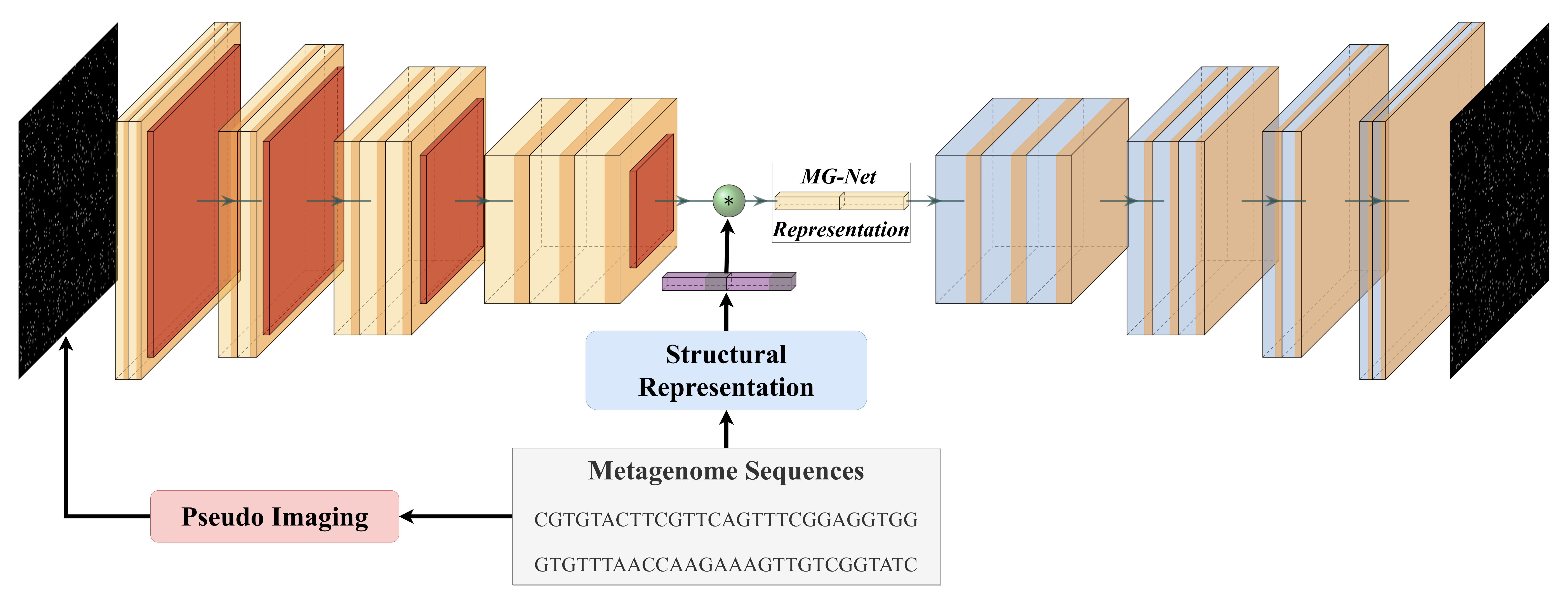}
    \caption{\textbf{Overall Architecture.} Our approach, MG-Net, is illustrated here. There are three major components (i) a global structural feature, (ii) a pseudo-imaging module to generate multi-modal representation, and (iii) integrated structural reasoning with attention for learning robust features with self-supervision.}
    \label{fig:arch}
\end{figure*}
One of the major challenges in metagenome-based diagnostics is the need for specialized bioinformatics pipelines for analysis of the enormous amounts of DNA sequences for detecting disease markers~\cite{chiu2019clinical,hasan2020metagenomics}. 
Machine learning research offers several opportunities to analyze DNA sequences collected directly from environment samples~\cite{ching2018opportunities}. 
Deep learning models, in particular, have been explored for representation learning from metagenome sequences for many associated tasks including, but not limited to: 
capturing simple nucleotide representations with reverse complement CNNs and LSTMs~\cite{bartoszewicz2020deepac}, depth-wise separable convolutions to predict taxonomy of metagenome sequences~\cite{busia2019deep}, genomic sub-compartment prediction~\cite{ashoor2020graph} and disease gene predictions~\cite{hwang2019humannet} with graph representations, predicting taxonomy of sequences by learning representations with bidirectional LSTMs with \emph{k}-mer embedding and self attention mechanism~\cite{liang2020deepmicrobes}, learning metagenome representations with ResNet~\cite{He_2016_CVPR} to predict the taxonomy. 

\textbf{Pseudo-Imaging} is often used in astrophysics~\cite{nelson2006pseudo} and medicine~\cite{pennec2003tracking} to study objects/tissues by forming images in another modality using alternative sensing methods, which exhibit detailed representations compared to conventional imaging. In particular, pseudo imaging is widely used in medicine to obtain pseudo-CT estimations from MRI images~\cite{leu2020generation} and ultrasound deformation fields~\cite{sun2018research}. 
Deep learning models are particularly suited to handling pseudo image data, due to the success of convolutional neural networks in computer vision research~\cite{He_2016_CVPR,simonyan2014very}. 
However, there have been very few methods used for metagenome analysis. 
For example, Self-Organizing Maps (SOM)~\cite{nguyen2019metagenome} and Growing Self-Organizing Maps (GSOM)~\cite{nguyen2020growing} have been used to represent metagenome sequences as images and a shallow CNN model was used for disease prediction. A matrix representation of a polygenetic tree has been used with CNN to predict host phenotype of metagenome sequences~\cite{reiman2020popphy}.

\section{MG-NET: Leveraging Pseudo-Imaging for Multimodal Metagenome Analysis}\label{sec:proposed}
In this section, we introduce our MG-NET framework for extracting robust, self-supervised representations from metagenome data. Our approach has three major components: (i) capturing a global structural prior for each metagenome sequence conditioned on the metagenome structure, (ii) extracting local structural features from pseudo-images generated from metagenome sequences, and (iii) integrate local and global structural features in an integrated, attention-driven structural reasoning module for multi-modal feature extraction. The overall approach is illustrated in Figure \ref{fig:arch}. 
We jointly model the global and local structural properties in a unified framework, which is trained in a self-supervised manner, \textit{without labels}, to capture robust representations aimed for metagenome classification with limited and unbalanced labeled data. 

\subsection{Capturing the Global Structure with Graph Representations}\label{sec:global}
First, we construct a global graph representation of the \textit{entire} metagenome sample, i.e., the graph provides a structural representation of the sequenced clinical sample. We take inspiration from the success of De Bruijn graphs for genome analysis~\cite{lin2016assembly,narayanan2020genome} and use a modified version to represent the metagenome sample. Given a metagenome sample $\mathcal{X_i}$ with sequence reads $X_0, X_1, \ldots X_n \in \mathcal{X_i}$, we construct a weighted, directed graph whose nodes are populated by k-mers $x_j$ such that $x_0, x_1, \ldots x_l \in X_i$. 
Each k-mer is a subsequence from a genome read $X_i$ of length $k$, extracted using a sliding window of length $k$ and stride $s$. 
Each edge direction is determined by order of occurrence of each observed k-mer in the sliding window. 
The edge weights are iteratively updated based on the observation of the co-occurrence of the nodes and are a function of the frequency of co-occurrence of the k-mers. 
The updated weights are given by
\begin{equation}
    \Psi(x_i, x_j) {=} f_s(\frac{e_{i,j}}{(\Vert e_{i,j} - e^{\prime}_{i,j} \Vert_2)})
    \label{eqn:global_weights}
\end{equation}
where $e_{i,j}$ is the current weight between the k-mer nodes $x_i$ and $x_j$ and $e^{\prime}_{i,j}$ is the new weight to be updated; $f_s(\cdot)$ is a weighted update function that bounds the new weight within a given range. In our experiments, we bound the edge weights to be between -2 and 2 and hence set $f_s(q) = 2 \sqrt{max(q - 1, 1)} + (min(q {-} 2, 2) {+} 2)$ to capture the \textit{relative} increase in frequency to highlight structures that emerge through repeated co-occurrence while suppressing spurious links. 
The edge weights are initially set to $1$.
Given this structural representation, we extract features for each k-mer using  \textit{node2vec}~\cite{grover2016node2vec} to capture the ``community'' or neighborhood structure of a k-mer to reject clutter due to observation noise~\cite{laver2015assessing}. 
The resulting representation $x^{st}_i$ for each k-mer $x_i$ captures its neighborhood within a sequence and provides a structural prior over the metagenome structure. The global structural representation for a sequence $X_i$ is the average-pooled (AP) representations of each k-mer given by $X^g_i{=}AP(\{x^{st}_1, x^{st}_2, \ldots x^{st}_n\})$. 

\begin{figure*}[t]
    \centering
    \begin{tabular}{cc}
        \includegraphics[width=0.34\textwidth]{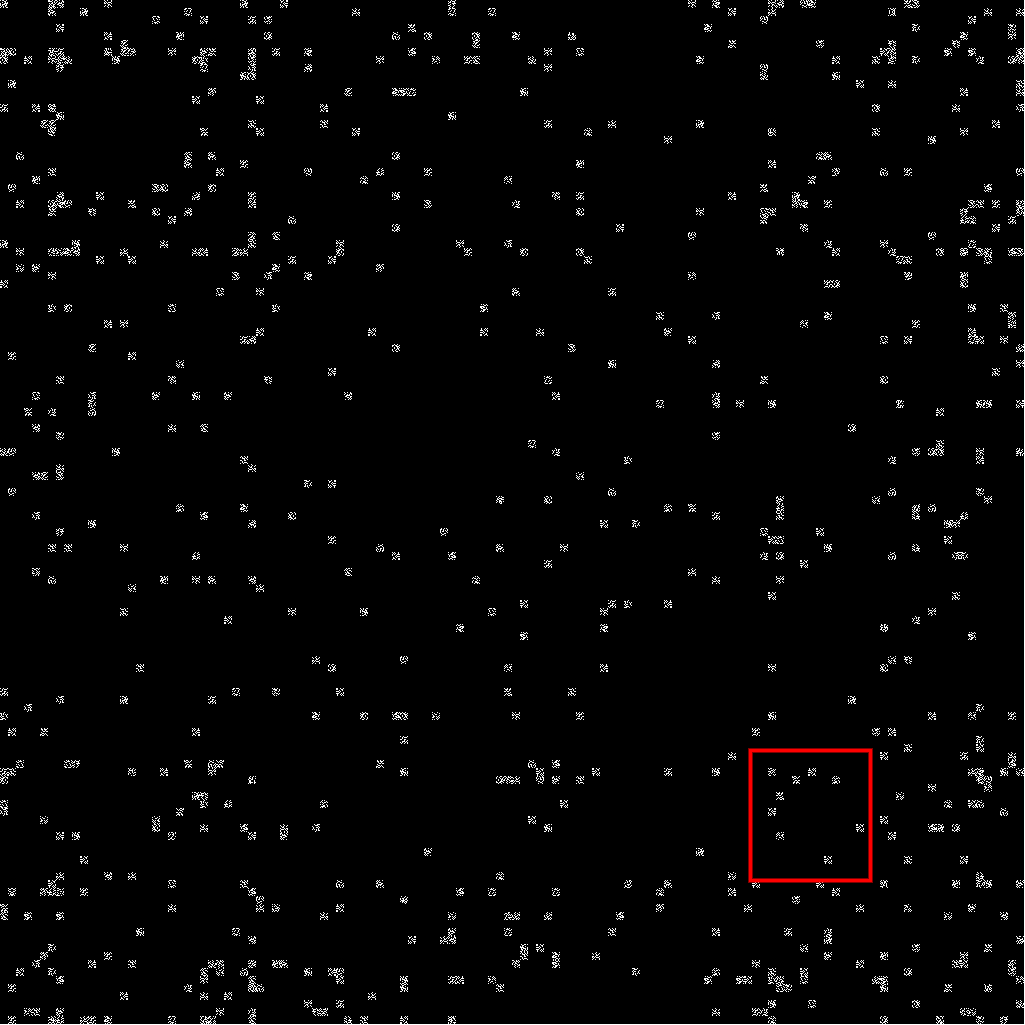} & \includegraphics[width=0.34\textwidth]{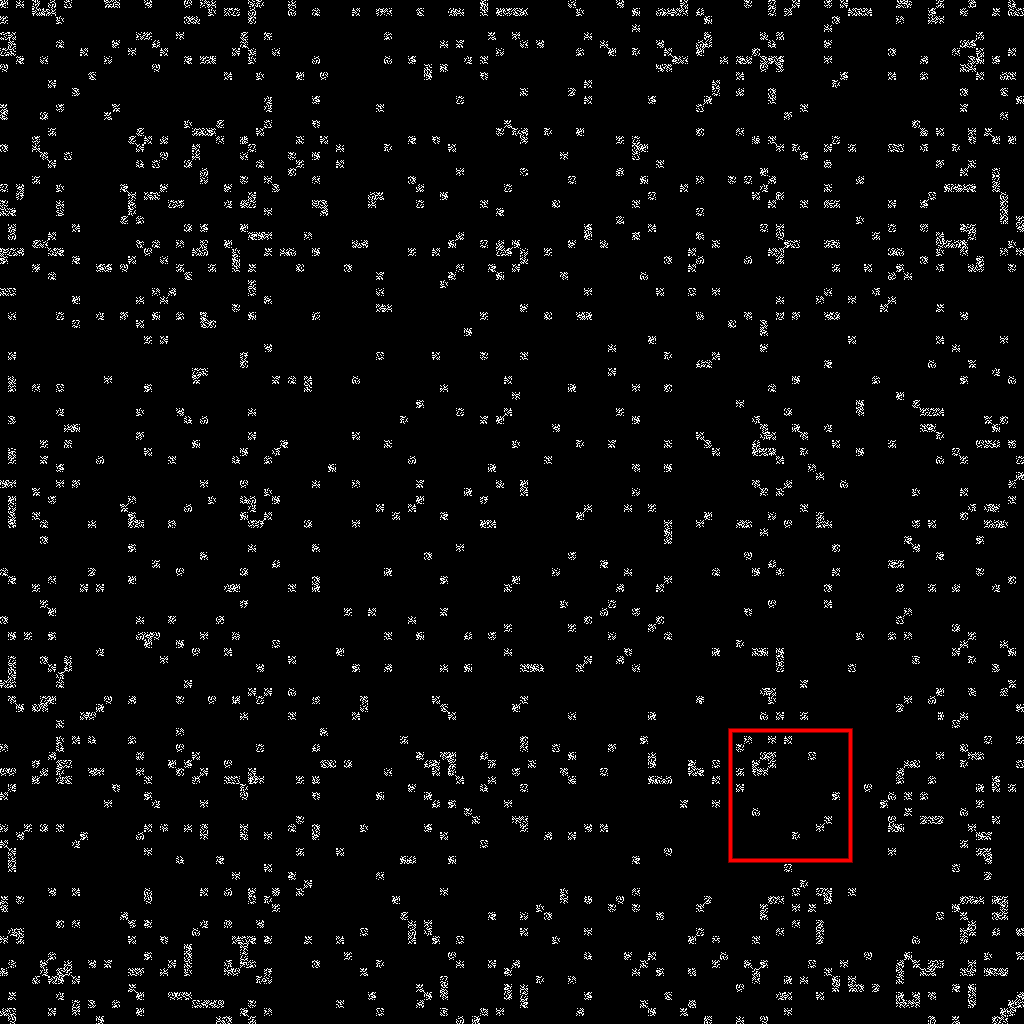} \\
        (a) & (b)
    \end{tabular}
    
    \caption{\textbf{Pseudo-Images} generated from our approach. Highly distinct patterns returned from attention maps for the species \textit{Mannheimia haemolytica} from different clinical samples are shown with a red bounding box. 
    }
    \label{fig:pseudo-image-eg}
\end{figure*}
\subsection{Pseudo-Imaging for Local Structural Properties}\label{sec:local}
The second step in our framework is to generate a pseudo-image $I_{r}$ for each metagenome sequence read $X_r\in\mathcal{X}$. The key intuition behind generating a pseudo-image is to represent and learn recurring patterns (or ``fingerprints''~\cite{stobbe2013probe}) in metagenome sequence reads that belong to the same pathogen species automatically. 
For example, in Figure~\ref{fig:pseudo-image-eg}, it can be seen that two sequences from the same pathogen \textit{Mannheimia haemolytica} have recurring patterns across clinical samples. 
Inspired by the success of Gray Level Co-occurrence Matrix (GLCM)~\cite{ballerini2009query,bagari2018combined}, we use a histogram-based formulation to provide a visual representation of a metagenome sequence. Instead of binary co-occurrence, we use \textit{relative} co-occurrence to generate images with varying intensity. 
Each ``\textit{pixel}'' $p_{i,j}\in I_r$ is representative of the frequency of co-occurrence between the k-mers $x_i$ and $x_j$ in a sequence read. The resulting pseudo-image representation is given by 
\begin{equation}
    I_r(i,j) = \sum_{i=1}^{4^k}\sum_{j=i+s}^{4^k} 
        \begin{cases}
            255*f(i,j)/N, & \text{if } f(x_i,x_j) > \lambda_{min} \\
            0, & \text{otherwise}
        \end{cases}
    \label{eqn:image_pixels}
\end{equation}
where $f(x_i, x_j)$ is the relative co-occurrence between k-mers $x_i$ and $x_j$ computed using Equation~\ref{eqn:global_weights}; $s$ is the stride length; $N$ is the sum of all co-occurrences to scale the value between $0$ and $1$;$\lambda_{min}$ is a cutoff parameter to reduce the impact of noise introduced due to any read errors~\cite{laver2015assessing}. Note that the $e(i,j)$ is computed \textit{per sequence read} and not at the sample level as done in Section~\ref{sec:global}. This allows us to model sequence-level patterns and hence capture species-specific patterns. The depth of the image is set to be $3$ and the pixel values are duplicated to simlate an RGB image and hence allow us to leverage advances in deep neural networks to extract automated features. In our experiments, we set $\lambda_{min}=0$. 

\subsection{Structural Reasoning with Attention}
The third and final step in our framework is to use the global representations (Section~\ref{sec:global}) to help learn local structural properties from pseudo-images (Section~\ref{sec:local}) using attention as a structural reasoning mechanism. Specifically, we use a convolutional neural network (CNN) to extract local structural features from a given pseudo-image. As can be seen from Figure~\ref{fig:arch}, we use the intermediate ($4^{th}$ convolutional block) layer of the CNN as a local feature representation $X^{lc}_i$ of a sequence read $X_i$. We obtain a robust representation by using attention-based reasoning mechanism given by
\begin{equation}
   X^{MG}_i = GAP(softmax({X^{g}_i\odot X^{lc}_i})\odot X^{g}_i) 
   \label{eqn:MGNET_Feat}
\end{equation}
where $GAP$ refers to the Global Average Pooling function~\cite{lin2013network} and $X^{g}_i$ is global structural feature representation provided by the global graph representation from Section~\eqref{sec:global}. We train the network end-to-end by having a decoder block (a mirrored network of deconvolutional operations) to reconstruct the input pseudo-image from this structural representation. Note that our goal is to learn robust representations from limited labeled metagenome data rather than reconstruction or segmentation. Hence, adding skip connections like U-Net~\cite{ronneberger2015u} will allow the network to “cheat” and not learn robust, “compressed” representations. We augment the features from the CNN with structural features from the node2vec representations using Equation~\ref{eqn:MGNET_Feat}, where we flatten the feature maps so that they match the dimensions for element-wise multiplication in the attention mechanism. We use L2-norm between the reconstructed and the actual pseudo-image as the objective function to train the network in a self-supervised manner. Formally, we define the loss function as $\mathcal{L}_{recons} = {\lVert I^\prime_{i} - I_{i}\lVert^{2}}$, where $I^\prime_{i}$ and $I_{i}$ refer to the reconstructed and actual pseudo-images, respectively. We show empirically (Section~\ref{sec:results}) that the integrated reasoning during training enhances the performance as opposed to mere concatenation of auto-encoder and global features. 

\subsection{Implementation Details}
We use a 4-layer convolutional neural network based on VGG-16~\cite{simonyan2014very} as our feature extractor in Section~\ref{sec:local}. We mirror the network to have a 4-layer decoder network to reconstruct the pseudo-image. The network is trained end-to-end for 25 epochs with a batch size of 64 and converges in about 30 minutes on a server with an NVIDIA Titan RTX and a 32-core AMD ThreadRipper CPU. The extracted representations are then finetuned for 10 epochs with a 3-layer deep neural network for pathogen classification. The learning rate is set to $1\times 10^{-4}$ for both stages and optimized using the standard gradient descent optimizer. Empirically, we find that having a k-mer length of $5$ and stride of $10$ provides the best results and present other variations in the ablation study (Section~\ref{sec:results}) for completeness. Default parameters were used for node2vec representations. All networks are trained from scratch during the pre-training phase.
\section{Experimental Evaluation}\label{sec:results}

\begin{table*}[t]
    \centering
    \resizebox{0.99\textwidth}{!}{
    \begin{tabular}{|c|c|c|c|c|c|c|c|c|c|c|c|c|c|c|}
    \toprule
    \multirow{1}{*}{\textbf{Max Training}} & \multicolumn{2}{|c|}{\textbf{Host}} & \multicolumn{2}{|c|}{\textbf{{H. somni.}}} & \multicolumn{2}{|c|}{\textbf{{M. bovis.}}} & \multicolumn{2}{|c|}{\textbf{{M. haemo.}}} & \multicolumn{2}{|c|}{\textbf{{P. multo.}}} & \multicolumn{2}{|c|}{\textbf{{T. pyoge.}}} & \multicolumn{2}{|c|}{\textbf{{B. treha.}}} \\
    \cline{2-15}
    \textbf{Sequences/Class} & Prec. & Rec. & Prec. & Rec. & Prec. & Rec. & Prec. & Rec. & Prec. & Rec. & Prec. & Rec. & Prec. & Rec.\\
    \toprule
    0 & 0.97 & 0.23 & 0.26 & 0.61 & 0.60 & 0.50 & 0.01 & 0.08 & 0.01 & 0.16 & 0.02 & 0.93 & 0.00 & 0.00  \\ 
    25 & 0.96 & 0.23 & 0.13 & 0.02 & 0.48 & 0.15 & 0.02 & 0.33 & 0.02 & 0.13 & 0.00 & 0.07 & 0.00 & 0.03 \\ 
    100 & 0.87 & 0.27 & 0.07 & 0.05 & 0.48 & 0.74 & 0.03 & 0.13 & 0.00 & 0.01 & 0.00 & 0.00 & 0.00 & 0.00 \\ 
    250 & 0.90 & 0.34 & 0.16 & 0.09 & 0.42 & 0.90 & 0.03 & 0.14 & 0.00 & 0.01 & 0.01 & 0.87 & 0.00 & 0.00 \\
    500 & 0.97 & 0.85 & 0.66 & 0.46 & 0.59 & 0.96 & 0.33 & 0.67 & 0.03 & 0.01 & 0.10 & 1.00 & 0.00 & 0.00 \\ 
    1000 & 0.98 & 0.89 & 0.70 & 0.57 & 0.68 & 0.96 & 0.34 & 0.61 & 0.04 & 0.01 & 0.14 & 1.00 & 0.00 & 0.10 \\ 
    \midrule
    MG-Net (All) & \textbf{0.99} & \textbf{0.98} & \textbf{0.84} & \textbf{0.74} & \textbf{0.88} & \textbf{0.95} & \textbf{0.59} & \textbf{0.69} & \textbf{0.42} & \textbf{0.18} & \textbf{0.94} & \textbf{1.00} & \textbf{0.02} & \textbf{0.33} \\ 
    \midrule
    \textit{Node2Vec}~\cite{narayanan2020genome} (All) & 0.87 & 0.65 & 0.16 & 0.39 & 0.19 & 0.14 & 0.04 & 0.24 & 0.05 & 0.01 & 0.02 & 0.13 & 0.00 & 0.00 \\ 
    
    \bottomrule
    \end{tabular}
    }
    \caption{\textbf{Recognition results.} Performance of the proposed MG-NET with varying number of training samples the recognition task on clinical metagenome data. Precision and Recall are reported for each class. Note that \textit{P. Multocida} and \textit{B. Trehalosi} have maximum of $21$ and $17$ samples across all settings, respectively.}
    \label{tab:recog}
\end{table*}
\textbf{Data Collection.} For constructing the dataset for the training and evaluation of automated metagenome-based pathogen detection, we collected metagenome sequences from $13$ Bovine Respiratory Disease Complex (BRDC) lung specimens at a local (name redacted to preserve anonymity) diagnostic laboratory using the DNeasy Blood and Tissue Kit (Qiagen, Hilden, Germany). 
Sequencing libraries are prepared from the extracted DNA using the Ligation Sequencing Kit and the Rapid Barcoding Kit. 
Prepared libraries are sequenced using MinION (R9.4 Flow cells), and sequences with an average Q-score of more than $7$ are used in the final genome. 
RScript MinIONQC~\cite{lanfear2019minionqc} was used for quality assessment. 

\textbf{Annotation and Quality Control.} 
We used the MiFi platform~\cite{espindola2021microbe}\footnote{https://bioinfo.okstate.edu/} for labeling metagenome sequence data. This platform is based on the modified version of the bioinformatics pipeline discussed by Stobbe \textit{et al.} \cite{stobbe2013probe}. Using MiFi, unique signature sequences referred to as \textit{e-probes} were developed for the pathogen of interest. These e-probes were then used to identify and label pathogen specific sequences in the metagenome reads, and differentiate them from other sequences (host, commensals, and other pathogen sequences). Clinical metagenome samples from $7$ patients were used for training, $1$ for validation, while sequences from $5$ patients were used for evaluation. 

\textbf{Metrics and Baselines.} 
To quantitatively evaluate our approach, we use precision, recall, and the F-score for each class. 
We do not use accuracy as a metric since real-life metagenomes can be highly skewed towards host sequences. Precision and recall, on the other hand, allow us to quantify the false alarm rates and provide more precise detection accuracy. 
We compare against other representation learning frameworks for metagenome analysis proposed, such as graph-based approaches~\cite{narayanan2020genome}, and a deep learning model termed Seq2Vec, an end-to-end sequence-based learning model based on DeePAC~\cite{bartoszewicz2020deepac}. For classification, we consider both traditional baselines such as logistic regression (LR), support vector machines (SVM), and multi-layer perceptron (MLP), as well as a deep neural network (DL). The MLP baseline has two hidden layers with 256 neurons each, while the deep learning baseline has 3 hidden layers with 256, 512 and 1024 neurons each with a ReLU activation function. 
We choose the hyperparameters for each of the baselines using an automated grid search and the best performing models from the validation set were taken for evaluation on the test set.

\begin{table*}[t]
    \centering
    \resizebox{0.99\textwidth}{!}{
    \begin{tabular}{|c|c|c|c|c|c|c|c|c|c|c|c|c|}
    \toprule
    \multirow{2}{*}{\textbf{Classifier}} & \multicolumn{2}{|c|}{\textbf{Node2Vec}~\cite{narayanan2020genome}} & \multicolumn{2}{|c|}{\textbf{SPK}~\cite{narayanan2020genome}} & \multicolumn{2}{|c|}{\textbf{WLK}~\cite{narayanan2020genome}} &
    \multicolumn{2}{|c|}{\textbf{GSK}~\cite{narayanan2020genome}} & \multicolumn{2}{|c|}{\textbf{Seq2Vec}} & \multicolumn{2}{|c|}{\textbf{MG-Net}}\\
    \cline{2-13}
    & \textbf{Host} & \textbf{Path.} & \textbf{Host} & \textbf{Path.} & \textbf{Host} & \textbf{Path.}  & \textbf{Host} & \textbf{Path.} & \textbf{Host} & \textbf{Path.} & \textbf{Host} & \textbf{Path.} \\
    \toprule
    Linear   & 0.05  & 0.07  & 0.02  & 0.04  & 0.04 & 0.05 & 0.03 & 0.04 & - & - & 0.86 & 0.38 \\
    LR       & 0.82  & 0.04  & 0.86  & 0.13  & 0.87 & 0.10 & 0.87 & 0.02 & - & - & 0.97 & 0.53 \\
    SVM      & 0.81  & 0.07 &  0.86  & 0.11  & 0.86 & 0.10  & 0.86 & 0.05 & - & - & 0.97  & 0.53 \\
    MLP      & 0.85  & 0.09  & 0.86  & 0.08  & 0.88  & 0.07  & 0.84 & 0.00 & - & - & \textbf{0.98}  & 0.54 \\
    DL       & 0.74  & 0.10  & 0.78  & 0.09  & 0.81 & 0.13  & 0.79 & 0.07 & 0.758 & 0.362 &  \textbf{0.98} & \textbf{0.63}\\
    \bottomrule
    \end{tabular}
    }
    \caption{
    \textbf{Comparison with other representations.} Performance evaluation of machine learning baselines using other metagenome representations. Average F1 scores are reported across pathogen and host classes. 
    }
    \label{tab:graph_rep_comp_all}
\end{table*}
\subsection{Quantitative Evaluation}
We evaluate our approach and0 report the quantitative results in Table~\ref{tab:recog} and Table~\ref{tab:graph_rep_comp_all}. We evaluate under different settings to assess the robustness of the proposed framework under limited data and limited supervision. We also compare against comparable representation learning approaches to highlight the importance of attention-based reasoning to integrate global and local structural information in a unified framework. We significantly outperform all baselines when training with the entire training data and offer competitive performance when fine-tuned with only $500$ labeled samples per class. 

\textbf{Effect of Limited Labels.} 
Since our representations are learned in a self-supervised manner, we also evaluate its metagenome recognition performance with limited labeled data and summarize results in Table~\ref{tab:recog}. First, we evaluate when there is no labeled data using k-means clustering to segment the features into groups and align the predicted clusters with the ground-truth using the Hungarian method, following prior works~\cite{ji2019invariant}. It can be seen that we perform reasonably well considering we do not use any labels from the ground-truth to train. As expected, the performance gets better with the use of increasing amounts of labeled data. It is interesting to note that we match the performance of fully supervised models like sequence-level graph-based representations~\cite{narayanan2020genome} and end-to-end deep learning models like Seq2Vec with as little as $500$ samples and outperform their performance with as little as $1000$ labeled samples per class. Given the performance of the linear classifier (Table~\ref{tab:graph_rep_comp_all}), we can see that our approach learns robust representations with limited data.

\begin{table}[t]
    \centering
    \begin{tabular}{|c|c|c|c|c|c|c|}
    \toprule
        \multirow{2}{*}{\textbf{Approach}} & \multicolumn{3}{|c|}{\textbf{Host}} & \multicolumn{3}{|c|}{\textbf{Pathogen}} \\
        \cline{2-7}
         & \textbf{Precision} & \textbf{Recall} & \textbf{F1} & \textbf{Precision} & \textbf{Recall} & \textbf{F1} \\
         \toprule
         Autoencoder Only & 0.950 & 0.800 & 0.869 & 0.242 & 0.453 & 0.315 \\
         Autoencoder + Structural Priors & 0.980 & 0.970 & 0.975 & 0.523 & 0.543 & 0.533 \\
         Structural Priors Only & 0.990 & 0.960 & 0.975 & 0.522 & 0.615 & 0.565\\
         \midrule
         MG-Net (k=3, s=5) & 0.980 & 0.980 & 0.980 & 0.598 & 0.628 & 0.613\\
         MG-Net (k=3, s=10) & 0.99 & 0.980 & 0.984 & 0.595 & 0.640 & 0.617\\
         MG-Net (k=4, s=5) & 0.980 & 0.980 & 0.980 & 0.610 & 0.598 & 0.604\\
         MG-Net (k=4, s=10) & 0.980 & 0.980 & 0.980 & 0.580 & 0.581 & 0.581\\
         MG-Net (k=5, s=5) & 0.980 & 0.980 & 0.980 & 0.573 & 0.588 & 0.581\\
         \midrule
         MG-Net$^*$ (k-5, s=10) & \textbf{0.990} & \textbf{0.980} & \textbf{0.984} & \textbf{0.615} & \textbf{0.648} & \textbf{0.631}\\
    \bottomrule
    
    \end{tabular}
    \caption{\textbf{Ablation Studies.} Performance evaluation of different variations to evaluate the effect of each design choice in the overall framework. $^*$ indicates final model.}
    \label{tab:ablation}
\end{table}

\textbf{Comparison with Other Representations.} We compare our representations with other baselines and summarize the results in Table~\ref{tab:graph_rep_comp_all}. We use a mix of traditional approaches (MLP, SVM, and LR) and a deep neural network (DL). We also train a linear classifier on top of each of the representations to assess their robustness. As can be seen, the representations from MG-Net outperform all other baselines by a significant margin. In fact, a linear classifier outperforms all other baseline representations that use deep neural networks. Our MG-Net features with a deep learning classifier achieve an average pathogen F-score of $63\%$ and a host F-score of $98\%$, which are significantly higher than baseline representations. Evaluation with 5-fold cross validation (see supplementary material) corroborate the results. 
It is interesting to note that both the graph kernels and Seq2Vec use sequence read-level features, and our approach with only an autoencoder from Table~\ref{tab:ablation} has comparable performance indicating that the global structural features have a significant impact on the performance.

\textbf{Ablation Studies.} Finally, we systematically evaluate each component of the framework independently to identify their contribution. Specifically, we provide ablations of our approach with using only image features (autoencoder only), image+structural features without the MG-Net architecture (Autoencoder + Structural Priors), and only structural features (from node2vec).
From Table~\ref{tab:ablation}, we can see that the use of structural priors greatly improves the performance. We remove the structural prior and use an autoencoder trained on only the pseudo-images (\textit{Autoencoder Only}) and use a late fusion strategy (\textit{Autoencoder + Structural Priors}) to evaluate the structural reasoning module. While the performance is better than other baselines, the final MG-Net architecture outperforms all variations. Finally, we also vary the length of k-mer sequences and stride lengths and see that the performance increases with an increase in stride lengths while reducing the k-mer length reduces the performance. 

\section{Conclusion and Future Work}
In this work, we presented MG-Net, one of the first efforts to offer a multi-modal perspective to metagenome analysis using the idea of co-occurrence statistics to construct pseudo-images. 
A novel, attention-based structural reasoning framework is introduced to perform multi-modal feature fusion, allowing for joint optimization over multiple modalities. Extensive real-world clinical data experiments show that the learned representations outperform existing baselines by a significant margin and offer a way forward for metagenome classification under limited resources. We aim to leverage these results to build automated diagnosis and intervention pipelines for novel pathogen diseases with limited supervision. 
\section{Acknowledgement}
This research was supported in part by the US Department of Agriculture (USDA) grants AP20VSD and B000C011.

We thank Dr. Kitty Cardwell and Dr. Andres Espindola (Institute of Biosecurity and Microbial Forensics, Oklahoma State University) for providing access and assisting with use of the MiFi platform. 
%
%
%
%
\bibliographystyle{splncs04}
\bibliography{egbib}

\begin{thebibliography}{10}
\providecommand{\url}[1]{\texttt{#1}}
\providecommand{\urlprefix}{URL }
\providecommand{\doi}[1]{https://doi.org/#1}

\bibitem{ashoor2020graph}
Ashoor, H., Chen, X., Rosikiewicz, W., Wang, J., Cheng, A., Wang, P., Ruan, Y.,
  Li, S.: Graph embedding and unsupervised learning predict genomic
  sub-compartments from hic chromatin interaction data. Nature communications
  \textbf{11}(1),  1--11 (2020)

\bibitem{bagari2018combined}
Bagari, A., Kumar, A., Kori, A., Khened, M., Krishnamurthi, G.: A combined
  radio-histological approach for classification of low grade gliomas. In:
  International MICCAI Brainlesion Workshop. pp. 416--427. Springer (2018)

\bibitem{ballerini2009query}
Ballerini, L., Li, X., Fisher, R.B., Rees, J.: A query-by-example content-based
  image retrieval system of non-melanoma skin lesions. In: MICCAI International
  Workshop on Medical Content-Based Retrieval for Clinical Decision Support.
  pp. 31--38. Springer (2009)

\bibitem{bartoszewicz2020deepac}
Bartoszewicz, J.M., Seidel, A., Rentzsch, R., Renard, B.Y.: Deepac: predicting
  pathogenic potential of novel dna with reverse-complement neural networks.
  Bioinformatics  \textbf{36}(1),  81--89 (2020)

\bibitem{busia2019deep}
Busia, A., Dahl, G.E., Fannjiang, C., Alexander, D.H., Dorfman, E., Poplin, R.,
  McLean, C.Y., Chang, P.C., DePristo, M.: A deep learning approach to pattern
  recognition for short dna sequences. BioRxiv p. 353474 (2019)

\bibitem{ching2018opportunities}
Ching, T., Himmelstein, D.S., Beaulieu-Jones, B.K., Kalinin, A.A., Do, B.T.,
  Way, G.P., Ferrero, E., Agapow, P.M., Zietz, M., Hoffman, M.M., et~al.:
  Opportunities and obstacles for deep learning in biology and medicine.
  Journal of The Royal Society Interface  \textbf{15}(141),  20170387 (2018)

\bibitem{chiu2019clinical}
Chiu, C.Y., Miller, S.A.: Clinical metagenomics. Nature Reviews Genetics
  \textbf{20}(6),  341--355 (2019)

\bibitem{espindola2021microbe}
Espindola, A.S., Cardwell, K.F.: Microbe finder (mifi{\textregistered}):
  Implementation of an interactive pathogen detection tool in metagenomic
  sequence data. Plants  \textbf{10}(2), ~250 (2021)

\bibitem{grover2016node2vec}
Grover, A., Leskovec, J.: node2vec: Scalable feature learning for networks. In:
  Proceedings of the 22nd ACM SIGKDD international conference on Knowledge
  discovery and data mining. pp. 855--864 (2016)

\bibitem{hamburg2010path}
Hamburg, M.A., Collins, F.S.: The path to personalized medicine. New England
  Journal of Medicine  \textbf{363}(4),  301--304 (2010)

\bibitem{hasan2020metagenomics}
Hasan, M.R., Sundararaju, S., Tang, P., Tsui, K.M., Lopez, A.P., Janahi, M.,
  Tan, R., Tilley, P.: A metagenomics-based diagnostic approach for central
  nervous system infections in hospital acute care setting. Scientific Reports
  \textbf{10}(1),  1--11 (2020)

\bibitem{He_2016_CVPR}
He, K., Zhang, X., Ren, S., Sun, J.: Deep residual learning for image
  recognition. In: Proceedings of the IEEE Conference on Computer Vision and
  Pattern Recognition (CVPR) (June 2016)

\bibitem{hwang2019humannet}
Hwang, S., Kim, C.Y., Yang, S., Kim, E., Hart, T., Marcotte, E.M., Lee, I.:
  Humannet v2: human gene networks for disease research. Nucleic acids research
   \textbf{47}(D1),  D573--D580 (2019)

\bibitem{ji2019invariant}
Ji, X., Henriques, J.F., Vedaldi, A.: Invariant information clustering for
  unsupervised image classification and segmentation. In: Proceedings of the
  IEEE/CVF International Conference on Computer Vision. pp. 9865--9874 (2019)

\bibitem{lanfear2019minionqc}
Lanfear, R., Schalamun, M., Kainer, D., Wang, W., Schwessinger, B.: Minionqc:
  fast and simple quality control for minion sequencing data. Bioinformatics
  \textbf{35}(3),  523--525 (2019)

\bibitem{laver2015assessing}
Laver, T., Harrison, J., O’neill, P., Moore, K., Farbos, A., Paszkiewicz, K.,
  Studholme, D.J.: Assessing the performance of the oxford nanopore
  technologies minion. Biomolecular detection and quantification  \textbf{3},
  ~1--8 (2015)

\bibitem{leu2020generation}
Leu, S.C., Huang, Z., Lin, Z.: Generation of pseudo-ct using high-degree
  polynomial regression on dual-contrast pelvic mri data. Scientific reports
  \textbf{10}(1),  1--11 (2020)

\bibitem{liang2020deepmicrobes}
Liang, Q., Bible, P.W., Liu, Y., Zou, B., Wei, L.: Deepmicrobes: taxonomic
  classification for metagenomics with deep learning. NAR Genomics and
  Bioinformatics  \textbf{2}(1),  lqaa009 (2020)

\bibitem{lin2013network}
Lin, M., Chen, Q., Yan, S.: Network in network. arXiv preprint arXiv:1312.4400
  (2013)

\bibitem{lin2016assembly}
Lin, Y., Yuan, J., Kolmogorov, M., Shen, M.W., Chaisson, M., Pevzner, P.A.:
  Assembly of long error-prone reads using de bruijn graphs. Proceedings of the
  National Academy of Sciences  \textbf{113}(52),  E8396--E8405 (2016)

\bibitem{metzker2010sequencing}
Metzker, M.L.: Sequencing technologies—the next generation. Nature reviews
  genetics  \textbf{11}(1),  31--46 (2010)

\bibitem{mikheyev2014first}
Mikheyev, A.S., Tin, M.M.: A first look at the oxford nanopore minion
  sequencer. Molecular ecology resources  \textbf{14}(6),  1097--1102 (2014)

\bibitem{narayanan2020genome}
Narayanan, S., Ramachandran, A., Aakur, S.N., Bagavathi, A.: Gradl: A framework
  for animal genome sequence classification with graph representations and deep
  learning. In: 2020 19th IEEE International Conference on Machine Learning and
  Applications (ICMLA). pp. 1297--1303. IEEE (2020)

\bibitem{nelson2006pseudo}
Nelson, R.J., Mooney, J.M., Ewing, W.S.: Pseudo imaging. In: Algorithms and
  Technologies for Multispectral, Hyperspectral, and Ultraspectral Imagery XII.
  vol.~6233, p. 62330M. International Society for Optics and Photonics (2006)

\bibitem{nguyen2020growing}
Nguyen, H.T., Nguyen, B.A., Nguyen, M.N., Truong, Q.D., Nguyen, L.C., Banh,
  T.T.N., Linh, P.D.: Growing self-organizing maps for metagenomic
  visualizations supporting disease classification. In: International
  Conference on Future Data and Security Engineering. pp. 151--166. Springer
  (2020)

\bibitem{nguyen2019metagenome}
Nguyen, T.H.: Metagenome-based disease classification with deep learning and
  visualizations based on self-organizing maps. In: International Conference on
  Future Data and Security Engineering. pp. 307--319. Springer (2019)

\bibitem{pennec2003tracking}
Pennec, X., Cachier, P., Ayache, N.: Tracking brain deformations in time
  sequences of 3d us images. Pattern Recognition Letters  \textbf{24}(4-5),
  801--813 (2003)

\bibitem{reiman2020popphy}
Reiman, D., Metwally, A.A., Sun, J., Dai, Y.: Popphy-cnn: a phylogenetic tree
  embedded architecture for convolutional neural networks to predict host
  phenotype from metagenomic data. IEEE journal of biomedical and health
  informatics  \textbf{24}(10),  2993--3001 (2020)

\bibitem{ronneberger2015u}
Ronneberger, O., Fischer, P., Brox, T.: U-net: Convolutional networks for
  biomedical image segmentation. In: International Conference on Medical image
  computing and computer-assisted intervention. pp. 234--241. Springer (2015)

\bibitem{simonyan2014very}
Simonyan, K., Zisserman, A.: Very deep convolutional networks for large-scale
  image recognition. arXiv preprint arXiv:1409.1556  (2014)

\bibitem{stobbe2013probe}
Stobbe, A.H., Daniels, J., Espindola, A.S., Verma, R., Melcher, U.,
  Ochoa-Corona, F., Garzon, C., Fletcher, J., Schneider, W.: E-probe diagnostic
  nucleic acid analysis (edna): a theoretical approach for handling of next
  generation sequencing data for diagnostics. Journal of microbiological
  methods  \textbf{94}(3),  356--366 (2013)

\bibitem{sun2018research}
Sun, H., Xie, K., Gao, L., Sui, J., Lin, T., Ni, X.: Research on pseudo-ct
  imaging technique based on an ultrasound deformation field with binary mask
  in radiotherapy. Medicine  \textbf{97}(38) (2018)

\end{thebibliography}
\end{document}